\documentclass[sigconf,natbib=true,anonymous=false]{acmart}
\usepackage{multirow}
\usepackage{xcolor}
\usepackage{bbm}
\usepackage[bottom]{footmisc}
\usepackage{float}
\usepackage{tikz}
\usepackage{xurl}
\usepackage{pgfplots}
\usepgfplotslibrary{groupplots}
\usepackage{graphics}
\usepackage{longtable}
\usepackage{caption}
\usepackage{hyperref}
\usepackage{subcaption}
\pgfplotsset{compat=1.18}
\AtBeginDocument{%
  \providecommand\BibTeX{{%
    \normalfont B\kern-0.5em{\scshape i\kern-0.25em b}\kern-0.8em\TeX}}}

\setcopyright{acmlicensed}
\copyrightyear{2018}
\acmYear{2018}
\acmDOI{XXXXXXX.XXXXXXX}

\acmConference[Conference acronym 'XX]{Make sure to enter the correct
  conference title from your rights confirmation emai}{June 03--05,
  2018}{Woodstock, NY}
\acmISBN{978-1-4503-XXXX-X/18/06}

\begin{document}

\title{Q-PEFT: Query-dependent Parameter Efficient Fine-tuning for Text Reranking with Large Language Models}

\author{Zhiyuan Peng}
\authornote{Both authors contributed equally to this research.}
\orcid{0000-0002-9870-4422}
\affiliation{%
  \institution{Santa Clara University}
  \city{Santa Clara}
  \country{USA}}
\email{zpeng@scu.edu}

\author{Xuyang Wu}
\authornotemark[1]
\orcid{0000-0002-8807-0016}
\affiliation{%
  \institution{Santa Clara University}
  \city{Santa Clara}
  \country{USA}}
\email{xwu5@scu.edu}

\author{Qifan Wang}
\orcid{0000-0002-7570-5756}
\affiliation{%
  \institution{Meta AI}
  \city{Menlo Park}
  \country{USA}}
\email{wqfcr@meta.com}

\author{Sravanthi Rajanala}
\orcid{0009-0009-1813-9995}
\affiliation{%
  \institution{Walmart Global Tech}
  \city{Sunnyvale}
  \country{USA}}
\email{sravanthi.rajanala@walmart.com}

\author{Yi Fang}
\authornote{Yi Fang is the corresponding author.}
\orcid{0000-0001-6572-4315}
\affiliation{%
  \institution{Santa Clara University}
  \city{Santa Clara}
  \country{USA}}
\email{yfang@scu.edu}
\renewcommand{\shortauthors}{Trovato and Tobin, et al.}

\begin{abstract}


Parameter Efficient Fine-Tuning (PEFT) methods have been extensively utilized in Large Language Models (LLMs) to improve the down-streaming tasks without the cost of fine-tuing the whole LLMs. Recent studies have shown how to effectively use PEFT for fine-tuning LLMs in ranking tasks with convincing performance; there are some limitations, including the learned prompt being fixed for different documents, overfitting to specific tasks, and low adaptation ability. In this paper, we propose a query-dependent parameter efficient fine-tuning (Q-PEFT) approach for text reranking, which provides LLMs with insights about the true queries, thereby facilitating the generation of true queries from input documents. Specifically, we utilize the query to extract the top-$k$ tokens from concatenated documents, serving as contextual clues. We further augment Q-PEFT by substituting the retrieval mechanism with a multi-head attention layer to achieve end-to-end training and cover all the tokens in the documents, guiding the LLMs to generate more document-specific synthetic queries, thereby further improving the reranking performance. Extensive experiments are conducted on four public datasets, demonstrating the effectiveness of our proposed approach.


\end{abstract}

\ccsdesc[500]{Information systems~Query reformulation}
\ccsdesc[500]{Information systems~Learning to rank}
\ccsdesc[500]{Computing methodologies~Natural language generation}

\keywords{Large Language Models, Query-dependent, Parameter Efficient Fine-tuning, Reranking}



\maketitle

\section{Introduction}

Recently, with the rapid development of large language models (LLMs), it has been increasingly recognized that leveraging their remarkable language understanding, generation, generalization, and reasoning abilities can significantly enhance the performance of many complex tasks \cite{DBLP:journals/tmlr/WeiTBRZBYBZMCHVLDF22}, such as natural language processing \cite{Radford2019LanguageMA, brown2020language, DBLP:journals/corr/abs-2302-13971}, recommendation systems \cite{DBLP:journals/corr/abs-2305-07001, DBLP:journals/corr/abs-2305-08845, DBLP:journals/corr/abs-2306-10933}, finance \cite{DBLP:journals/corr/abs-2303-17564}, and medicine \cite{DBLP:journals/corr/abs-2311-05112}. In the field of Information Retrieval (IR), models have dynamically evolved from original term-based methods to integration with advanced neural models \cite{DBLP:conf/emnlp/KarpukhinOMLWEC20, khattab2020colbert}. Although IR neural models enhance the original term-based models by capturing complex semantic signals, there are still numerous challenges, such as data scarcity, interpretability, generation, and model fairness, that current IR systems must overcome to provide more accurate and robust services. A lot of work has been done deploying LLMs in novel IR systems, which improve overall performance in aspects like query understanding, query rewriting, retrieval, and reranking. LLMs not only extend the retrieval and reranking capabilities of traditional IR models but also exhibit stronger adaptability and compatibility in unknown tasks or data. Although leveraging LLMs to enhance IR systems has shown promising results in many areas, there are still many challenges associated with the use of LLMs. Original LLMs are primarily trained on publicly available text data, with models designed to predict the next word by calculating the generative likelihood of word sequences based on the contextual information from preceding words. Therefore, the foundational LLM models are more focused on text generation capabilities, rather than the strong retrieval and reranking abilities needed in information retrieval models \cite{DBLP:journals/tmlr/WeiTBRZBYBZMCHVLDF22}. For example, when leveraging LLMs for ranking a list of documents related to a query based on textual similarity, only GPT-4 achieves promising results, while other LLMs fail \cite{DBLP:conf/emnlp/0001YMWRCYR23, DBLP:journals/corr/abs-2305-02156, DBLP:journals/corr/abs-2310-07712}. This outcome is even inferior to traditional neural rankers. Even when given two documents and using LLMs to determine which document is more relevant to a given query, LLMs suffer from position bias issues \cite{DBLP:journals/corr/abs-2306-17563}. 

To better leverage the capabilities of LLMs for IR tasks, we need to optimize the basic models, which involves fine-tuning LLMs on task-specific ranking datasets to enable them to measure query-document relevance and fully understand ranking tasks \cite{DBLP:journals/corr/abs-1910-14424, DBLP:conf/sigir/JuYW21, DBLP:journals/corr/abs-2101-05667, DBLP:journals/corr/abs-2310-08319}. Fine-tuning the entire model has achieved significant improvements for traditional IR models, but this approach is not easily implemented in LLMs. First, LLMs require more data, especially high-quality and diverse data \cite{DBLP:conf/nips/BrownMRSKDNSSAA20}. Original LLMs are based on massive amounts of data to converge and improve performance in specific tasks while ranking tasks often lack large training data samples \cite{DBLP:journals/tmlr/WeiTBRZBYBZMCHVLDF22}. Second, even with abundant data, the whole model fine-tuning process of LLMs requires substantial computational resources, which means that fine-tuning the entire model is very expensive \cite{DBLP:journals/corr/abs-2302-13971}. Third, due to the large number of parameters in LLMs, fine-tuning the entire model poses significant risks. During the fine-tuning process, the model can easily lose its existing capabilities \cite{DBLP:journals/corr/abs-2309-01219}. To address these challenges, researchers have proposed more efficient parameter fine-tuning methods, especially for LLMs, which aim to guide LLM performance in new tasks by adjusting a small number of model parameters \cite{DBLP:conf/iclr/HuSWALWWC22}, adding external modules \cite{DBLP:conf/icml/HoulsbyGJMLGAG19} or soft prompt-based fine-tuning \cite{DBLP:conf/acl/LiL20, DBLP:journals/corr/abs-2103-10385, DBLP:conf/emnlp/LesterAC21}.

Some existing works in IR have utilized parameter-efficient fine-tuning (PEFT) methods to enhance the ranking ability of LLMs \cite{DBLP:conf/emnlp/TamLJXLLDT23,DBLP:conf/www/JungCR22, DBLP:conf/ecir/PalLDC23, DBLP:journals/corr/abs-2307-08303, DBLP:journals/csur/LiuYFJHN23, DBLP:conf/acl/ChoJSP23}. However, there are still limitations to these approaches. Some works employ LoRA to make the model more task-specific for ranking \cite{DBLP:conf/www/JungCR22}, but localized task-specific parameter adjustments also limit the model's generalization ability to adapt to other similar tasks \cite{DBLP:journals/corr/abs-2309-09055}. For soft-prompt based models \cite{DBLP:journals/corr/abs-2307-08303, DBLP:journals/csur/LiuYFJHN23, DBLP:conf/acl/ChoJSP23}, they typically learn a task-specific prompt representation, which can guide the model to generate task-specific outputs. However, the prompt module becomes fixed after tuning, making it inflexible for different tasks. Adapter-based models may be more flexible for replacement \cite{DBLP:conf/ecir/PalLDC23}, but existing study \cite{DBLP:conf/cikm/MaGZFC22} claim that they are not reliable across different tasks. In this work, we present two novel query-dependent parameter-efficient fine-tuning methods: Q-PEFT-R and Q-PEFT-A for text reranking. In our approach, we do not fine-tune the whole model or use LoRA for partial parameter tuning. Instead, we add a new query-dependent module, which is flexible and can be easily attached to or detached from the original LLM like adapter. In addition, it is superior to soft prompt-based methods because it is query-specific; the module's enhancement is based on each query-document pair, making it more generalized for different ranking tasks. Indeed, during model training, we freeze the original LLM and only fine-tune the parameters on the query-dependent module. Therefore, we can maintain the model's foundational capabilities while simultaneously endowing it with enhanced ranking abilities, achieving this with only minimal adjustments to the number of parameters. The Q-PEFT-R method leverages the top $k$ query-dependent words in a document, while the Q-PEFT-A method uses attention scores to weigh document-query correlations. Both methods guide the LLM to generate more document-specific synthetic queries, thereby affecting the ranking relevance score between the query and documents. The fundamental idea behind these methods is to use a query-dependent module to enhance the weighting between queries and similar documents and to diminish the weighting between queries and irrelevant documents. This approach aims to better guide the LLM in text ranking tasks. Our main contributions can be summarized as follows:

\begin{itemize}

\item To the best of our knowledge, this is the first work to learn a query-dependent module in Parameter Efficient Fine-Tuning (PEFT) for text reranking.

\item We propose two Q-PEFT methods Q-PEFT-R and Q-PEFT-A to guide the model to generate more document-specific synthetic queries, thereby affecting the ranking relevance score between the query and documents.

\item A comprehensive set of experiments are conducted on four widely used public datasets. The results demonstrate that our approach further improves over the baseline retrievers and significantly outperforms the hard prompting approach.

\item By building on a fully open-source LLM, we not only ensure reproducible and deterministic experimental results, but also anticipate exciting opportunities for integrating LLMs into end-to-end information access applications. We will make the code and checkpoints publicly available upon paper acceptance.

\end{itemize}

\section{Related Work}
\subsection{Reranking with LLMs}

Reranker, as a second-stage document filter in IR systems, aims to reorder the list of documents retrieved by a retriever (such as BM25) based on query-document relevance. Existing reranking methods using LLMs can be categorized into two paradigms: supervised reranker and unsupervised reranker \cite{DBLP:journals/corr/abs-2308-07107}. Supervised reranker methods \cite{DBLP:journals/corr/abs-1910-14424, DBLP:conf/sigir/JuYW21, DBLP:journals/corr/abs-2101-05667, DBLP:journals/corr/abs-2310-08319} involve fine-tuning LLMs on task-specific ranking datasets to address the inability of pre-trained LLMs to adequately measure query-document relevance and fully understanding reranking tasks. MonoBERT \cite{DBLP:journals/corr/abs-1910-14424} based on the pre-trained model BERT \cite{DBLP:conf/naacl/DevlinCLT19}, utilizes query-document pairs to enhance the relevance score based on the [cls] token. T5 \cite{DBLP:conf/emnlp/NogueiraJPL20}, along with \cite{DBLP:conf/sigir/JuYW21} and \cite{DBLP:journals/corr/abs-2101-05667}, can be fine-tuned to generate classification tokens “true” or “false” for relevant or irrelevant query-document pairs. RankLLaMa \cite{DBLP:journals/corr/abs-2310-08319}, relying on a decode-only structure, uses the last token representation [EOS] for relevance calculation. These methods typically employ small-scale LLMs with a promising performance. As LLMs increase in size, recent efforts \cite{DBLP:journals/corr/abs-2211-09110, DBLP:journals/corr/abs-2310-14122, DBLP:conf/emnlp/SachanLJAYPZ22, DBLP:conf/emnlp/SachanLJAYPZ22, DBLP:conf/emnlp/Zhuang0KZ23} have focused on prompting LLMs to directly enhance document reranking in an unsupervised manner. Based on pointwise methods, \citet{DBLP:journals/corr/abs-2211-09110} uses an LLM to generate binary labels “yes” or “no” to indicate whether a document is relevant to the query, as instructed by the prompt and query-document, with the relevance score calculated based on the log-likelihood of the tokens “yes” and “no”. \citet{DBLP:journals/corr/abs-2310-14122} introduces fine-grained relevance labels such as “highly relevant,” “somewhat relevant,” and “not relevant”, enabling LLMs to more effectively differentiate among documents with varying relevance levels to a query. \citet{DBLP:conf/emnlp/SachanLJAYPZ22} and \citet{DBLP:conf/emnlp/Zhuang0KZ23} propose a zero-shot document reranking approach based on the query generation method, where the query-document relevance score is generated by the average log-likelihood of producing the actual query tokens based on the document. From listwise perspective, some methods \cite{DBLP:conf/emnlp/0001YMWRCYR23, DBLP:journals/corr/abs-2305-02156, DBLP:journals/corr/abs-2310-07712} attempt to direct LLMs to output the reranked document list directly. However, the experiments show that only the GPT-4 based method achieves competitive performance, and it is highly sensitive to the document order in the prompt. For the pairwise method, \citet{DBLP:journals/corr/abs-2306-17563} attempts to instruct the generation of a high document identifier based on a document pair and the query. Nevertheless, pairwise methods still face high time complexity, as they require comparing relevance among all document pairs. Although the recent studies of using LLMs for document reranking have achieved significant progress, several challenges still persist. The time complexity is lower for the pointwise method, but the reranking performance has not reached the expected level. The listwise method has achieved convincing performance on GPT-4, but it is costly and operates as a "black box." For the pairwise method, although it can achieve performance comparable to the listwise method, its lower efficiency is a major limiting factor.


\subsection{Parameter Efficient Fine-tuning}

Parameter-efficient fine-tuning (PEFT) presents an effective solution by reducing the number of parameters required for fine-tuning and memory usage, while still achieving performance comparable to full fine-tuning \cite{DBLP:journals/corr/abs-2312-12148}. One direction in PEFT is low-rank decomposition, hypothesizing that weight changes during model optimization have a low intrinsic rank. For example, LoRA \cite{DBLP:conf/iclr/HuSWALWWC22} trains rank decomposition matrices (a down-project and an up-project) for the dense layer to approximate weight updates. On the other hand, addition-based methods introduce extra parameters by inserting small neural modules such as adapters \cite{DBLP:conf/icml/HoulsbyGJMLGAG19} or soft prompt-based fine-tuning \cite{DBLP:conf/acl/LiL20, DBLP:journals/corr/abs-2103-10385, DBLP:conf/emnlp/LesterAC21}. In these methods, only the additional parameters are tuned while the original parameters of pre-trained models remain frozen. The adapter method was first introduced in multi-domain image classification \cite{DBLP:conf/nips/RebuffiBV17}, facilitating efficient knowledge transfer across multiple visual domains. Soft prompt-based fine-tuning involves appending soft prompts or prefix vectors to input embeddings or hidden states during fine-tuning. Prompt tuning \cite{DBLP:conf/emnlp/LesterAC21} incorporates additional learnable prompt tokens into the model input, where only the prompt parameters are updated through gradient descent during fine-tuning. Prefix-tuning \cite{DBLP:conf/acl/LiL20} extends prompt-tuning by pretending trainable prefix (token) vectors to the keys and values of the self-attention layers. P-tuning \cite{DBLP:journals/corr/abs-2103-10385} represents another advancement in this field.

With significant progress in PEFT methods in LLMs, these methods have also been widely deployed in information retrieval tasks. \citet{DBLP:conf/emnlp/TamLJXLLDT23} demonstrates that parameter-efficient learning in neural text retrieval can achieve comparable performance to full-parameter fine-tuning in an in-domain setting, while drastically reducing parameter usage. \citet{DBLP:conf/www/JungCR22} first apply prefix-tuning and LoRA to the re-ranking stage, finding that these two kinds of parameter-efficient methods can outperform full fine-tuning on small test data. Adapters-SPLADE \cite{DBLP:conf/ecir/PalLDC23} examines adapters for sparse retrieval models, deploying adapters for SPLADE to retain efficiency and effectiveness, further improving SPLADE method performance. \citet{DBLP:conf/coling/LitschkoVG22} proposes Adapters and Sparse Fine-Tuning Masks, decoupling language-specific and task-specific knowledge for a more lightweight and effective zero-shot transfer to multilingual and cross-lingual retrieval tasks. SPTAR \cite{DBLP:journals/corr/abs-2307-08303} introduces soft prompt tuning, optimizing only the prompts’ embedding layer during training, allowing LLMs to better adapt to generating pseudo-queries and striking a balance between training cost and generation quality. DPTDR \cite{DBLP:journals/csur/LiuYFJHN23} applies prompt-tuning to improve dual encoders for Embedding-based Retrieval Models. Co-prompt \cite{DBLP:conf/acl/ChoJSP23} is proposed for better prompt generation in reranking tasks, while PaRaDe \cite{DBLP:conf/emnlp/DrozdovZD0RWAIM23} introduces a difficulty-based method to select few-show demonstrations for inclusion in the prompt, showing significant improvements compared to zero-shot prompting. Although many studies leverage PEFT methods to enhance model performance, as noted by \cite{DBLP:conf/cikm/MaGZFC22}, naively applying existing parameter-efficient fine-tuning methods from NLP literature may result in limited success in retrieval applications.

\begin{figure*}[t]
\centering
\includegraphics[width=1.0\textwidth]{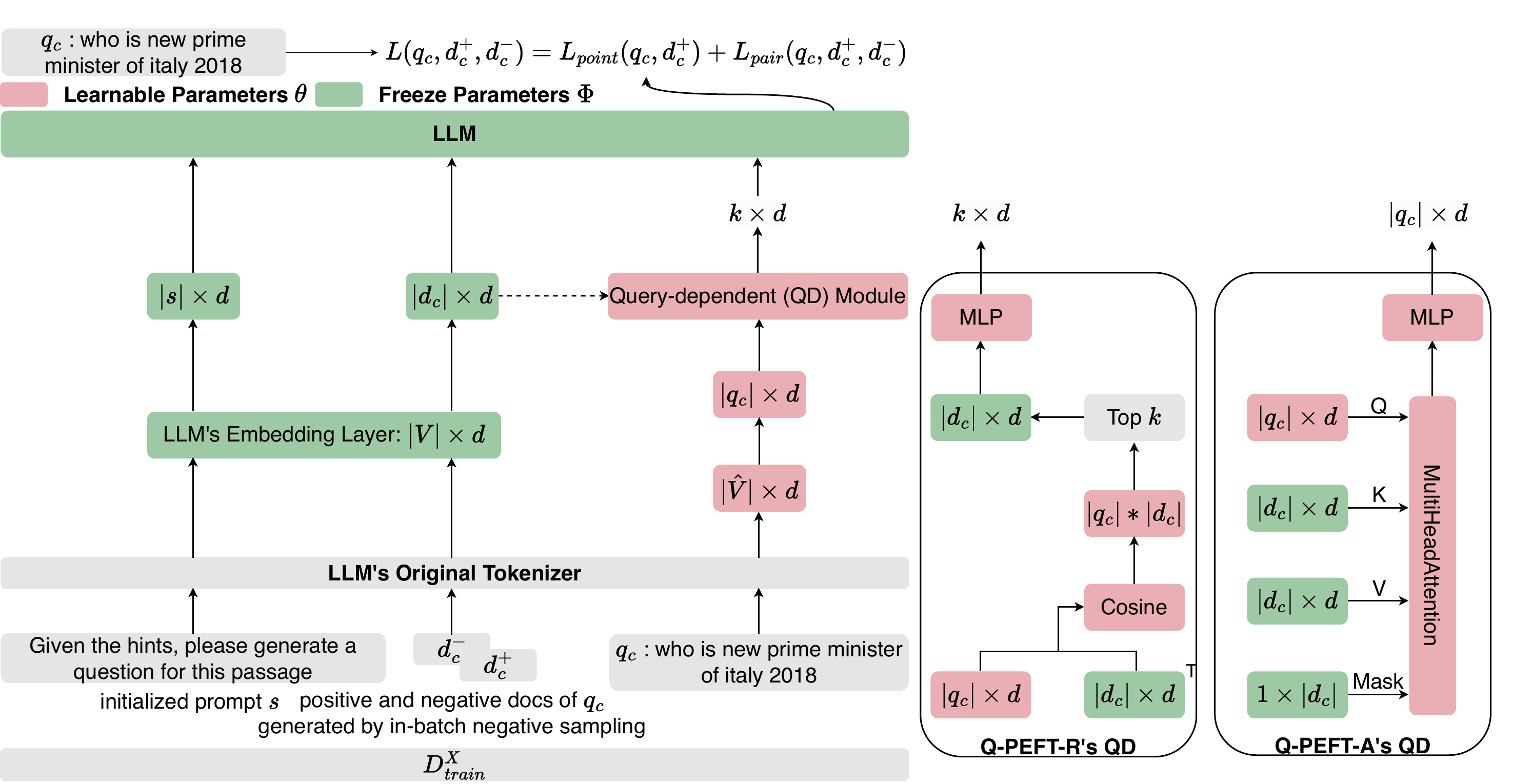}
\caption{The architecture of the proposed Q-PEFT. For each instance $\left\langle q_c, d_c^{+}, d_{c}^{-}\right\rangle$, we compute pointwise loss $L_{point}$ and pairwise loss $L_{pair}$ and add them together as final loss $L$. During training, LLM's original parameters $\Phi$ are fixed, and only the parameters of Query-dependent module $\theta$ are updated.}
\label{fig: q-peft}
\end{figure*}

\section{Parameter Efficient Query-aware Text Ranker}
Our research focuses on reranking the documents retrieved by retrievers. Instead of letting the LLMs guess what the true query looks like only depending on the input document, Q-PEFT leaks the information of the true query to LLMs to make the guessing or generation process much easier. Specifically, we utilize a query-dependent (QD) module to learn an embedding matrix and then append it with the embeddings of the prompt and input document, and finally, the concatenated embeddings are inputted into LLMs to compute the log-likelihood of the true query conditioned on the prompt and the input document as the query-document relevant score. Based on the different QD strategies, our Q-PEFT methods have two variations: Q-PEFT-R and Q-PEFT-A, the QD modules of which are explained in Section \ref{sec: q-peft-r} and Section \ref{sec: q-peft-a} respectively. In this section, we will first introduce the general Q-PEFT framework and then illustrate the two QD strategies in detail.

\subsection{Q-PEFT}
Assume a dataset $D=\left\{\left\langle q_i, d_i^{+}, d_{i, 1}^{-}, \cdots, d_{i, n}^{-}\right\rangle\right\}_{i=1}^m$ consists of $m$ instances, where each instance includes one positive document $d_i^{+}$ and a list of $n$ negative documents $d_{i, 1}^{-}, \cdots, d_{i, n}^{-}$. Dataset $D$ is split into $D_{train}$, $D_{eval}$ and $D_{test}$. We sample $X$ instances from $D_{train}$ as $D_{train}^{X}$ and similarly $D_{eval}^{Y}$ is sampled from $D_{eval}$. Q-PEFT is fine-tuned on $D_{train}^{X}$ and evaluated on $D_{eval}^{Y}$ to learn a query-dependent embedding matrix. At the training and evaluation processes, for each batch data $D_{batch}$ of $b$ queries, we randomly sample one negative document from $d_{i, 1}^{-}, \cdots, d_{i, n}^{-}$ for each query $q_i$ and then generate more instances by in-batch negative sampling. Specifically, for query $q_{i}$, its in-batch negative documents are $\{d_{j, j\neq i}^{+}\}_{j=1}^{b}$ and all the negative documents $\{d_j^{-}\}_{j=1}^{b}$ in the same batch $d$. After in-batch negative sampling, for each query $q$, it has $2b-1$ instances in which the only positive document is repeated $2b-1$ times to be paired with each negative document. Finally, we have $D_{batch} = \left\{\left\langle q_c, d_c^{+}, d_{c}^{-}\right\rangle\right\}_{c=1}^{b\times(2b-1)}$.

Figure \ref{fig: q-peft} illustrates the architecture of the proposed Q-PEFT. As shown in Figure \ref{fig: q-peft}, prompt $s$ is initialized as ``\textit{given the hints, please generate a query for this passage}'' the length of which is $l_{s}$. $f_{\theta}(\cdot)$ denotes the learnable parameters, including an embedding layer $f_{embed}(\cdot)$ initialized by the weights of LLM's original embedding layer and the parameters of the QD module $f_{QD}(\cdot)$. $f_{\theta}(q)$ represents the learned embedding matrix which contains the information of the true query. The green blocks represent LLM's original parameters $\Phi$ that are retained during the training. After training, $f_{\theta^{*}}(q)$ represents the learned embedding matrix where $\theta^{*}$ denotes the optimized $\theta$. The log-likelihood of query $q$ conditioned on document $d$ and prompt $s$ is defined as:

\begin{equation}
\begin{aligned}
I_{\theta, \Phi}(q| d, s)= \sum_{l=1}^{|q|} \log p_{\theta, \Phi}\left(q_{l} \mid q_{<l}, d, s\right)
\end{aligned}
\label{eq: log-likelihood}
\end{equation}
where $p_{\theta, \Phi}\left(q_{l} \mid q_{<l}, d, s\right)$ represents the possibility of predicting the current token by looking at previous tokens. Pointwise loss is applied to constrain the model to output a ground-truth-like query based on the input positive or relevant document, which is defined on one instance $\left\langle q_c, d_c^{+}, d_{c}^{-}\right\rangle$ as:

\begin{equation}
\begin{aligned}
L_{point}(q_{c}, d_{c}^{+})= & - I_{\theta, \Phi}(q_{c}| d_{c}^{+}, s)
\end{aligned}
\label{eq: poinwise}
\end{equation}

Pointwise loss is order sensitive, for instance, query ``\textit{who is new prime minister of Italy 2018}'' should equal ``\textit{2018 who is Italy's new prime minister}'', but obviously the pointwise losses are different for the two queries conditioned on the same positive document. To loosen this strict loss function $L_{point}$, we add a pairwise loss $L_{pair}$ as the regularization under the assumption that the query generated by the negative document should be less relevant to the ground-truth query otherwise it should be punished. Also, Q-PEFT can learn from negative documents with this $L_{pair}$. Inspired by hinge loss, on one instance $\left\langle q_c, d_c^{+}, d_{c}^{-}\right\rangle$, pairwise loss $L_{pair}$ is defined as:


\begin{equation}
\begin{aligned}
L_{pair}(q_{c}, d_{c}^{+}, d_{c}^{-}) = \max \left\{0, I_{\theta, \Phi}(q_{c}| d_{c}^{-}, s) - I_{\theta, \Phi}(q_{c}| d_{c}^{+}, s) \right\}
\end{aligned}
\label{eq: pairwise}
\end{equation}
Our final loss $L$ directly combines the pointwise and pairwise losses:
\begin{equation}
\begin{aligned}
L(q_{c}, d_{c}^{+}, d_{c}^{-})= L_{point}(q_{c}, d_{c}^{+}) + L_{pair}(q_{c}, d_{c}^{+}, d_{c}^{-})
\end{aligned}
\label{eq: our}
\end{equation}


During the inferencing process, the Q-PEFT model reranks the top-$k$ documents, denoted as ${z_{1}, z_{2}, ... z_{K}}$, which are retrieved by the retriever $R$. The relevance score for this reranking is based on the log-likelihood of the generated question $q$ conditioned on each document $z_{j}$ along with the prompt $s$. This is expressed as $I_{\theta^{*}, \Phi}(q| z_{j}, s)$, where $s$ represents the initialized prompt and $\theta^{*}$ denotes the learned optimal parameters after training phase.

\subsection{QD Module of Q-PEFT-R} \label{sec: q-peft-r}
Q-PEFT-R directly utilizes the true query to retrieve the top-$k$ similarity tokens in the document, converting the retrieved tokens into embeddings by the learnable embedding layer initialized by the weights of LLM's original embedding layer. Then, a multi-layer perception (MLP) is applied to learn the embedding matrix further. 

More formally, for query $q_{c}$, the learnable embedding layer $f_{embed}$ with shape $\hat{V} \times d$ where $|V|$ is LLM's vocabulary size, and $d$ is the LLM's dimensionality. $f_{embed}$ converts the tokenized $|q_{c}|$ ids into embedding matrix $|Q_{c}| \times d$. we compute the cosine similarity for each token in $q_{c}$ with each token in document $d_{c}$ by:

\begin{equation}
\begin{aligned}
cos(q_{c}, d_{c}) = \frac{f_{embed}(q_{c}) \times LLM_{embed}(d_{c})^{T}}{|f_{embed}(q_{c})| \times |LLM_{embed}(d_{c})|}
\end{aligned}
\end{equation}
where $LLM_{embed}(\cdot)$ represents LLM's original embedding layer, which is frozen during the training and $cos(q_{c}, d_{c})$ has shape $|q_{c}| \times |d_{c}|$. We choose top-$k$ unique tokens in $d_{c}$ by the cosine similarity scores and then find their corresponding embedding in $d_{c} \times d$ to get the learnable embedding matrix $k \times d$. The last MLP doesn't change the shape of the learned embedding matrix. 

\subsection{QD Module of Q-PEFT-A} \label{sec: q-peft-a}

Even though Q-PEFT-R shows improvements over baseline models in our experiments, it has some disadvantages. First, based on cosine similarity, Q-PEFT-R may oversimplify the representation by focusing only on the most similar tokens. It might miss out on the broader context or less obvious but relevant tokens, leading to a less effective representation. Second, important information might be spread in long documents, and limiting the selection to the top-$k$ tokens might miss crucial details. Third, cosine similarity is a relatively simple measure and may not capture more complex relationships or dependencies between words and phrases. Last but not least, selecting the top-$k$ tokens is not inherently differentiable, making it not an end-to-end training model. 

To fix the shortness of Q-PEFT-R, we propose to apply a multi-head attention layer to replace the retrieval process in Q-PEFT-R. Multi-head attention inherently considers the entire context of the query and the document. It evaluates the relevance of each word in its specific context, leading to a more nuanced understanding of the text. The attention mechanism can weigh all parts of the document, regardless of length, allowing it to capture relevant information even from distant parts of the text. Multi-head attention can learn a variety of complex patterns and dependencies, as each head can potentially focus on different types of relationships. What's more, the attention mechanism is fully differentiable, making it well-suited for gradient-based optimization and end-to-end training in neural network models. Specifically, we treat query $q_{c}$ as parameter $Q$ of the multi-head attention layer, and $K$ and $V$ parameters are both the document $d_{c}$ to learn an embedding matrix with shape $|q_{c}| \times d$. 

\section{Experimental Setup}

\subsection{Retriever} \label{sec: retriever}

Given a collection $C$ consisting of $M$ documents ${d_{1}, d_{2}, ... d_{M}}$, a retriever $R$ retrieves the top-$k$ documents ${z_{1}, z_{2}, ... z_{k}}$ for query $q$. Q-PEFT can work with any retriever. In this paper, we select a combination of unsupervised retriever methods, which include BM25 \cite{DBLP:journals/ftir/RobertsonZ09}, MSS \cite{DBLP:conf/acl/SachanPSKPHC20}, and Contriever \cite{DBLP:journals/tmlr/IzacardCHRBJG22}, as well as supervised retrievers in the form of DPR \cite{DBLP:conf/emnlp/KarpukhinOMLWEC20} and MSS-DPR \cite{DBLP:conf/acl/SachanPSKPHC20} to enable a thorough evaluation of Q-PEFT's performance across different retrieval paradigms:

\begin{itemize}
\item \textbf{BM25} is commonly employed to evaluate the relevance between search terms and documents, making it a cornerstone in the field of IR. Numerous previous studies, as evidenced by recent work \cite{DBLP:journals/corr/abs-2104-05740}, have demonstrated that BM25 remains a robust baseline method. 
\item \textbf{MSS} represents a dense retriever that is trained in a joint manner with the retriever and reader components. This joint training is accomplished by predicting masked salient spans within the text, such as named entities. 
\item \textbf{Contriever} is a self-supervised model employing a contrastive learning framework for pre-training in IR. It demonstrates competitive performance when compared to BM25.
\item \textbf{DPR} represents a dense passage retriever founded on the bi-encoder architecture. It is initialized with a BERT-based network and undergoes discriminative training using query-document relevance pairs. During training, challenging negatives are sampled, which include top passages returned by BM25 that do not contain the answer but match most question tokens.
\item \textbf{MSS-DPR} represents a notable advancement in enhancing the performance of DPR through a two-step process: first, by pre-training the dense retriever using the MSS (Masked Salient Spans) method, and subsequently, by subjecting it to DPR-style supervised fine-tuning.
\end{itemize}

\subsection{Datasets}

Following the existing work, our study utilized several widely used public reranking datasets including Natural Questions (NQ) \cite{DBLP:journals/tacl/KwiatkowskiPRCP19}, TriviaQA \cite{DBLP:conf/acl/JoshiCWZ17}, SQuAD \cite{DBLP:conf/emnlp/RajpurkarZLL16} and WebQ \cite{DBLP:conf/emnlp/BerantCFL13}. We employed the training and development datasets made available by DPR \cite{DBLP:conf/emnlp/KarpukhinOMLWEC20}, integrating the top-$k$ results obtained from multiple retrievers except BM25 released by UPR \cite{DBLP:journals/tmlr/IzacardCHRBJG22}. This is because the test queries for BM25 in UPR \cite{DBLP:conf/acl/SachanPSKPHC20} differed from those used with other retrievers and for consistent evaluation, we regenerated the test queries and top-$k$ passage retrieved by BM25 utilizing BEIR toolkit \cite{DBLP:conf/nips/Thakur0RSG21}. The evidence passages can be traced back to the English Wikipedia dump of December 2018. In alignment with the established DPR protocol, Wikipedia articles are partitioned into passages with each containing precisely 100 words. The statistics of our dataset is presented in Table \ref{tab: data}.

\begin{table}[t!]
\centering
\begin{tabular}{c|ccccccc}
\toprule
Dataset           & Train & Ret.Train & Eval & Test &\\
\hline
Natural Questions & 79,168 & 58,880 & 8,757 & 3,610 \\
TriviaQA          & 78,785 & 60,413 & 8,837 & 11,313\\
SQuAD             & 78,713 & 70,096 & 8,886 & 10,570\\
WebQ              & 3,417  & 2,474  & 361   & 2,032 \\
\bottomrule
\end{tabular}
\caption{The statistics of the datasets. The column \textbf{Train} refers to original training examples in the dataset. \textbf{Ret.Train} refers to the actual questions used for training supervised retrievers after filtering.}
\label{tab: data}
\end{table}

\subsection{Baseline Models and Evaluation Metrics}

\begin{table*}[!htbp]
\centering
\resizebox{\linewidth}{!}{%
\begin{tabular}{c|cc|cc|cc|cc|cc|cc}
\toprule
\multirow{2}{*}{ Retriever } & \multicolumn{2}{c}{ NQ } & \multicolumn{2}{c}{ WebQ } & \multicolumn{2}{c}{ SQuAD } &  \multicolumn{2}{c}{TriviaQA} & \multicolumn{2}{c}{Average} & \multicolumn{2}{c}{Avg Imp \%}\\
\cmidrule(lr){2-3} \cmidrule(lr){4-5} \cmidrule(lr){6-7} \cmidrule(lr){8-9} \cmidrule(lr){10-11} \cmidrule(lr){12-13}
& R@10 & H@10 & R@10 & H@10 & R@10 & H@10 & R@10 & H@10 & R@10 & H@10 & R@10 & H@10\\
\midrule
\multicolumn{13}{c}{ Unsupervised Retrievers } \\
\midrule
MSS             &19.19 &51.27 & 11.75  & 39.12  & 20.28   & 42.51   & 19.97    & 60.52    & 17.80 & 48.36 & - & -\\
+UPR            &33.71 &63.91 & 27.94  & 56.45  & 38.22   & 60.14   & 37.44    & 72.29    & 34.33 & 63.20 & 92.88\% & 30.69\%\\
+UPR-Inst       &33.35 &63.19 & 29.04  & 56.89  & 37.77   & 59.76   & 38.01    & 72.70    & 34.54 & 63.14 & 94.09\% & 30.57\%\\
+Q-PEFT-R          &$\dagger \ddagger$ 38.51 &$\dagger \ddagger$ 66.95 &$\dagger \ddagger$  29.70  &$\dagger \ddagger$  57.63  &$\dagger \ddagger$  \textbf{41.32}   &$\dagger \ddagger$  \textbf{61.71}   &$\dagger \ddagger$  43.44    &$\dagger \ddagger$  74.42    & 38.24 & \underline{65.18} & 114.88\% & \underline{34.79\%} \\
+Q-PEFT-A        &$\dagger \ddagger$ \textbf{38.96} &$\dagger \ddagger$  \textbf{67.20} &$\dagger \ddagger$  \textbf{30.38}  &$\dagger \ddagger$  \textbf{57.68}  &$\dagger \ddagger$  40.37   &$\dagger \ddagger$  61.24   &$\dagger \ddagger$  \textbf{44.08}    &$\dagger \ddagger$  \textbf{74.44}    & \underline{38.45} & 65.14 & \underline{116.03\%} & 34.71\% \\
\midrule
BM25            &22.01 &49.94 & 16.50  & 50.25  & 26.33   & 49.67   & 22.85    & 62.77    & 21.92 & 53.16 & - & -\\ 
+UPR            &32.31 &59.45 & 26.09  & 61.07  & 42.33   & 64.78   & 36.17    & 71.80    & 34.23 & 64.28 & 56.12\% & 20.91\%\\
+UPR-Inst       &31.75 &58.86 & 27.19  & 61.81  & 42.04   & 64.65   & 36.37    & 71.59    & 34.34 & 64.23 & 56.63\% & 20.82\%\\ 
+Q-PEFT-R          &$\dagger \ddagger$ \textbf{37.03} &$\dagger \ddagger$ \textbf{62.44} &$\dagger \ddagger$  \textbf{27.58}  &$\dagger \ddagger$  \textbf{62.01}  &$\dagger \ddagger$  44.98   &$\dagger \ddagger$  66.40   &$\dagger \ddagger$  41.95    &$\dagger \ddagger$  \textbf{73.67}    & 37.89 & \underline{66.13} & 72.81\% & \underline{24.40\%}\\
+Q-PEFT-A        &$\dagger \ddagger$ 36.56 &$\dagger \ddagger$ 61.69 &$\dagger \ddagger$  \textbf{27.58}  &$\dagger \ddagger$  61.91  &$\dagger \ddagger$  \textbf{46.05}   &$\dagger \ddagger$  \textbf{66.66}   &$\dagger \ddagger$  \textbf{42.56}    &$\dagger \ddagger$  73.65    & \underline{38.19} & 65.98 & \underline{74.19\%} & 24.12\%\\
\midrule
Contriever      &22.31 &58.73 & 14.83  & 56.40  & 26.06   & 54.65   & 20.27    & 68.00    & 20.87 & 59.45 & - & - \\
+UPR            &32.38 &67.12 & 21.99  & 64.12  & 41.25   & 69.06   & 32.58    & 75.86    & 32.05 & 69.04 & 53.59\% & 16.14\%\\ 
+UPR-Inst       &31.27 &66.07 & 22.91  & \textbf{64.47}  & 40.75   & 68.74   & 32.90    & 75.74    & 31.96 & 68.76 & 53.14\% & 15.66\%\\
+Q-PEFT-R          &$\dagger \ddagger$ 35.94 &$\dagger \ddagger$ 69.14 &$\dagger \ddagger$  22.86  &  $\dagger$ 63.29 &$\dagger \ddagger$  44.06   &$\dagger \ddagger$  70.90   &$\dagger \ddagger$  38.13    &$\dagger \ddagger$  77.58    & 35.25 & 70.23 & 68.91\% & 18.14\%\\
+Q-PEFT-A        &$\dagger \ddagger$ \textbf{37.52} &$\dagger \ddagger$ \textbf{70.47} &$\dagger \ddagger$  \textbf{23.75}  &$\dagger$  63.34  &$\dagger \ddagger$  \textbf{45.64}   &$\dagger \ddagger$  \textbf{71.80}   &$\dagger \ddagger$  \textbf{38.68}    &$\dagger \ddagger$  \textbf{77.88}    & \underline{36.40} & \underline{70.87} & \underline{74.42\%} & \underline{19.22\%}\\
\midrule
\multicolumn{13}{c}{ Supervised Retrievers } \\
\midrule
DPR             &38.74 &74.54 & 24.46  & 70.62  & 25.68   & 51.42   & 27.93    & 76.50    & 29.20 & 68.27 &  -&-\\    
+UPR            &41.73 &75.60 & 28.75  & 72.10  & 41.57   & 66.01   & 36.83    & 80.28    & 37.22 & 73.50 & 27.45\% & 7.66\%\\
+UPR-Inst       &40.28 &74.46 & 29.22  & \textbf{72.39}  & 41.51   & 65.94   & 36.88    & 80.19    & 36.97 & 73.25 & 26.61\% & 7.29\%\\
+Q-PEFT-R          &$\dagger \ddagger$ 44.95 &$\dagger \ddagger$ 77.09 &$\dagger \ddagger$  \textbf{29.37}  &$\dagger$  72.29  &$\dagger \ddagger$  44.34   &$\dagger \ddagger$  67.82   &$\dagger \ddagger$  41.47    &$\dagger \ddagger$  81.30    & 40.03 & 74.63 & 37.09\% & 9.31\%\\
+Q-PEFT-A        &$\dagger \ddagger$ \textbf{46.97} &$\dagger \ddagger$ \textbf{78.28} &$\dagger \ddagger$  29.25  &$\dagger$  71.95  &$\dagger \ddagger$  \textbf{45.14}   &$\dagger \ddagger$  \textbf{68.03}   &$\dagger \ddagger$  \textbf{42.41}    &$\dagger \ddagger$  \textbf{81.41}    & \underline{40.94} & \underline{74.92} & \underline{40.20\%} & \underline{9.74\%}\\
\midrule
MSS-DPR         &37.47 &77.48 & 21.75  & 71.65  & 33.25   & 65.85   & 25.54    & 79.15    & 29.50 & 73.53 & - & -\\
+UPR            &38.79 &76.81 & 22.78  & 72.59  & 46.96   & 77.10   & 31.33    & 81.56    & 34.97 & 77.02 & 18.52\% & 4.74\%\\
+UPR-Inst       &37.16 &75.35 & 23.14  & \textbf{72.93}  & 46.88   & 76.93   & 31.44    & 81.68    & 34.66 & 76.72 & 17.46\% & 4.34\%\\
+Q-PEFT-R          &$\dagger \ddagger$42.25 &$\dagger \ddagger$78.39 &$\dagger \ddagger$  23.01  &  71.65  &$\dagger \ddagger$  49.66   &$\dagger \ddagger$  78.68   &$\dagger \ddagger$  35.23    &$\dagger \ddagger$  82.51    & 37.54 & 77.81 & 27.23\% & 5.81\%\\
+Q-PEFT-A        &$\dagger \ddagger$ \textbf{44.37} &$\dagger \ddagger$ \textbf{80.30} &$\dagger \ddagger$  \textbf{23.34}  &  71.51  &$\dagger \ddagger$  \textbf{50.09}   &$\dagger \ddagger$  \textbf{78.82}   &$\dagger \ddagger$  \textbf{36.04}    &  \textbf{$\dagger \ddagger$82.62}    & \underline{38.46} & \underline{78.31} & \underline{30.36\%} & \underline{6.50\%}\\
\bottomrule
\end{tabular}
}
\caption{Comparative analysis of Top-10 Recall and Hit Ratio (R@10 and H@10) on test datasets, pre and post of UPR and Q-PEFT-based reranking models to the Top 100 retrieved passages. The optimal results for each retriever are highlighted in bold. Underlining indicates the best average metrics and average improvement across each retriever. The symbols $\dagger$ and $\ddagger$ denote statistically significant enhancements over the basic retrievers and UPR approach, respectively, with p-values < 0.01, as determined by a two-tailed t-test.}
\label{tab: main}
\end{table*}

As a competitive baseline model, we have chosen the Unsupervised Passage Retrieval (UPR) approach. UPR is a highly effective re-ranking method designed to enhance passage retrieval in open-question-answering scenarios. It leverages a pre-trained language model to calculate the probability of the input question conditioned on a retrieved passage. Notably, UPR can be applied in conjunction with various retrievers. In our study, we have replaced the pre-trained language model in UPR with Llama-2 to explore its potential and performance as well as for a fair comparison. In addition, leveraging the power of instruct examples, which can guide LLMs to generate more reasonable results. We utilize this technique to extend the performance of the UPR baseline. For the instruct-based UPR model, we randomly select three query and corresponding positive document pairs as instructive examples. Then, we run the UPR separately for each dataset to identify the most effective instructive example. The best query-document pair is subsequently used as the instructive example during model evaluation.

In our work, we employed Llama-2 \cite{DBLP:journals/corr/abs-2307-09288}, a family of LLMs that encompasses a spectrum of scales, spanning from 7 billion to 70 billion parameters. Specifically, for our study, we utilized the fine-tuned Llama-2-Chat model with 7 billion parameters \footnote{\url{https://huggingface.co/meta-llama/Llama-2-7b-chat-hf}}. This model had undergone fine-tuning processes, leveraging publicly available instruction datasets and benefiting from the input of over 1 million human annotations to optimize its performance.


We employed the top-$k$ Hit Ratio (H@$k$) to evaluate the reranking performance, ensuring a fair comparison with the UPR approach. This metric calculates the fraction of questions where at least one passage among the top-$k$ passages contains a span matching the human-annotated answer for that question. It is equivalent to the top-$k$ retrieval accuracy metric used in UPR \cite{DBLP:journals/tmlr/IzacardCHRBJG22}. In addition, we also employed the top-$k$ Recall (R@$k$) to measure the reranking performance, an evaluation metric that is widely used in IR models.

\section{Experimental Results}

\subsection{Implementation Details}

The model training and testing were conducted on Nvidia A100 GPUs with data type bfloat16 \cite{DBLP:journals/corr/abs-1905-12322} to expedite the training and testing processes. The training batch size is fixed at 4, and the inference batch size is fixed at 6. We conducted a maximum of 20 fine-tuning epochs and applied early stopping if there is no higher evaluation metric after 5 epochs. During training, we employed the Adam optimizer \cite{DBLP:journals/corr/KingmaB14}, and adopt in-batch negative sampling. We tried learning rate from \{1e-4, 3e-4, 5e-4, 7e-4, 9e-4, 1e-3, 3e-3, 5e-3, 7e-3, 9e-3, 1e-2, 3e-2, 5e-2, 7e-2, 9e-2, 1e-1\} and find 3e-2 is the best. We used the base configuration for the baseline models as specified in each respective paper. For our Q-PEFT model, the QD module was implemented based on a public package PEFT \cite{peft}. For Q-PFFT-R, we tried $k=5, 10, 15, 20$ and the number of MLP layers from 0 to 2. The best parameters for Q-PEFT-R are $k=10$ and one-layer MLP. For Q-PEFT-A, we tried the number of MLP layers from 0 to 2 and the number of attention heads from 2, 4, and 8. The best parameters for Q-PEFT-R are 2 attention heads and one-layer MLP.

\begin{table*}[!htbp]
\centering
\begin{tabular}{c|cc|cc|cc|cc|cc|cc}
\toprule
\multirow{2}{*}{ LLM } & \multicolumn{2}{c}{ Trainable } & \multicolumn{2}{c}{ BM25 } & \multicolumn{2}{c}{ MSS } & \multicolumn{2}{c}{ Contriever } & \multicolumn{2}{c}{ DPR } &  \multicolumn{2}{c}{ MSS-DPR } \\
\cmidrule(lr){2-3}\cmidrule(lr){4-5} \cmidrule(lr){6-7} \cmidrule(lr){8-9} \cmidrule(lr){10-11} \cmidrule(lr){12-13}
& Params & Pecentage & R@10 & H@10 & R@10 & H@10 & R@10 & H@10 & R@10 & H@10 & R@10 & H@10\\
\midrule
Llama-2-7b-chat & 231M & 3.32\% & 32.09 & 57.67 & 35.65  & 62.33  & 33.99 & 66.33 & 45.44 & 76.67 & 41.48 & 78.00\\
Llama-2-13b-chat & 321M & 2.41\% & \textbf{33.10} & \textbf{59.33} & \textbf{36.28}  & 63.33  & \textbf{36.07} & \textbf{68.33} & 46.35 & 78.00    & \textbf{44.01} & 78.33 \\
vicuna-7b-v1.5-16k & 231M & 3.32\% & 31.06 & 56.67 & 33.23 & 61.00 & 31.85 & 64.00 & 43.34 & 76.33 & 39.73 & 77.00\\
Mistral-7B-Instruct-v0.2 & 231M & 3.10\% & 32.53 & 57.67 & 35.99 & \textbf{63.67} & 35.26 & 67.00 & \textbf{46.53} & \textbf{78.33} & 43.61 & \textbf{80.33} \\
Mixtral\_7Bx2\_MoE & 231M & 1.77\% & 14.00 & 42.33 & 15.94  & 47.67 & 12.94 & 42.67 & 20.61 & 54.67 & 18.48 & 57.67\\
rank\_zephyr\_7b\_v1\_full & 231M & 3.10\% & 28.56 & 56.33 & 30.84 & 60.33 & 30.54 & 63.33 & 42.25 & 74.33 & 36.98 & 75.33\\
\bottomrule
\end{tabular}
\caption{Performance comparison of various large language models on BM25, MSS, Contriver, DPR, and MSS-DPR, evaluated on the NQ Dataset with Recall (R@10) and Hit Rate (H@10) Scores. Each model enumerates the trainable parameters and their corresponding percentage ratios relative to the total parameters. The best results for each retriever are highlighted in bold.}
\label{tab: llm_comparison}
\end{table*}

\subsection{Research Questions}
An extensive set of experiments was designed to address the following research questions:

\textbf{RQ1}: Can the proposed Q-PEFT framework achieve improved performance on reranking tasks over the baseline models? (Section~\ref{sec: main_results})

\textbf{RQ2}: How does the different LLMs influence the reranking effectiveness of our proposed model? (Section~\ref{sec: various_llms})

\textbf{RQ3}: How do different prompt designs, especially incorporating document hints, influence the performance and stability of outputs across various datasets and retrievers? (Section~\ref{sec: prompt})

\textbf{RQ4}: How does training size impact the performance of the Q-PEFT-R and Q-PEFT-A algorithms in reranking tasks across different Datasets? (Section~\ref{sec: training_size})

\subsection{Main Results}
\label{sec: main_results}

Our experiments follow the traditional two-stage retrieval and reranking tasks commonly used in Information Retrieval (IR). We evaluate the performance of baseline and Q-PEFT-based models on four different test datasets, focusing on Recall and Hit Rate metrics from the top 100 results of each query, based on various retrievers. Table \ref{tab: main} provides a detailed comparison of our proposed Q-PEFT-based model with baseline models across various datasets and based on different retrievers. The optimal results for each retriever are highlighted in bold. From the experimental results on each dataset, it is clear that our model demonstrates significant improvements based on each corresponding retriever model and UPR baseline model. Underlining indicates the best average metrics and average improvement across each retriever.

Comparing unsupervised and supervised retrievers, we can clearly see that our proposed reranking model exhibits more stable effects with unsupervised retrievers, with the best results of each dataset all coming from our proposed model. Similarly, the average improvement percentage across four different datasets based on our proposed models and basic retriever ranges from 74.19\% to 116.03\% on Recall@10, and from 19.22\% to 34.71\% on Hit Rate@10. These results indicate that our model, in the reranking process, not only consistently improves the overall ranking of all relevant documents but also enhances the ranking order of the top relevant documents in the top-ranking positions. On the other hand, when using supervised retrievers, which leverage fully fine-tuned retrievers on labeled data, our method can also enhance ranking performance. Although the percentage increase in performance of our proposed model compared to baseline models is not as high as unsupervised results, our proposed model is capable of further improving the best ranking performance of supervised retrievers. This is evident in the average Recall@10 across four different datasets, which increased from 38.45, the best result with an unsupervised retriever, to 40.94, the best result with a supervised retriever. Despite a slightly lower performance of our proposed model compared to the UPR baseline model in Hit Rate@10 on the Webq dataset, a comparison of the overall average results across all datasets shows that our proposed method still achieves significant growth. For example, the average Hit Rate@10 comparison between UPR-based and Q-PEFT-based models increased from the best 73.50 to 74.92 on a DPR-based retriever, and from the best 77.02 to 78.31 on an MSS-DPR based retriever. These results further demonstrate that our proposed model, whether based on unsupervised or supervised retrievers, can consistently enhance ranking performance.

\subsection{Q-PEFT Across LLMs}
\label{sec: various_llms}

Q-PEFT's effectiveness relies on the generative power of LLMs. Our goal is to validate its consistent performance across different LLM frameworks by evaluating it from three perspectives. First, we examine if increased LLM parameters, typically enhancing performance, similarly boost our method's effectiveness. Second, we assess its efficacy on various Llama 2 variants with fixed parameters. Third, we compare its performance in re-ranking-based LLMs using well-known LLM checkpoints from the Hugging Face platform.


\begin{itemize}
    \item Llama-2-13b-chat \footnote{\url{https://huggingface.co/meta-llama/Llama-2-13b-chat-hf}}: An upgraded version of the Llama-2-7b model, Llama-2-13b-chat has more parameters and demonstrates improved performance in many benchmark evaluations compared to Llama-2-7b.
    
    \item Vicuna-7b-v1.5-16k \footnote{\url{https://huggingface.co/lmsys/vicuna-7b-v1.5-16k}}: Vicuna is a chat assistant trained by fine-tuning Llama 2 on user-shared conversations collected from ShareGPT.
    
    \item Mistral-7B-Instruct-v0.2 \footnote{\url{https://huggingface.co/mistralai/Mistral-7B-Instruct-v0.2}}: An instruct fine-tuned version of the Mistral-7B-v0.1 generative text model, utilizing a variety of publicly available conversation datasets.

    \item Mixtral\_7Bx2\_MoE \footnote{\url{https://huggingface.co/cloudyu/Mixtral_7Bx2_MoE}}: A pretrained generative Sparse Mixture of Experts model, based on NurtureAI/neural-chat-7b-v3-16k and mncai/mistral-7b-dpo-v6.

    \item Rank\_zephyr\_7b\_v1\_full \footnote{\url{https://huggingface.co/castorini/rank_zephyr_7b_v1_full}}: A 7B parameter GPT-like model from \cite{DBLP:journals/corr/abs-2312-02724}, initially fine-tuned on a mix of publicly available, synthetic datasets, followed by task-specific listwise reranking data.
\end{itemize}

Table \ref{tab: llm_comparison} shows the performance of various language models and retrievers evaluated on the NQ dataset, measured by Recall@10 and Hit Rate@10. We fixed the model training hyper-parameters to ensure experimental fairness and minimize the impact of varying parameters on model performance. Firstly, a comparison between the Llama-2-7b-chat and Llama-2-13b-chat models clearly demonstrates that our proposed method's reranking capability improves as the model's parameter keeps increasing, further indicating the stability of our method. However, when comparing with Mixtral\_7Bx2\_MoE model, despite the increase in parameter number, there is a noticeable decrease in reranking effectiveness across all retrievers. This may be due to the fact that MOE architecture LLMs require more customized fine-tuning. A horizontal comparison of Llama-2-based models with 7B parameters reveals that the Mistral-7B-Instruct-v0.2 model outperforms the original Llama-2-7b-chat, with some results even comparable to the Llama-2-13b-chat's performance. Additionally, although the Vicuna-7b-v1.5-16k model is also based on the Llama-2 model structure, it does not perform as well in reranking tasks. Lastly, when comparing the rank\_zephyr\_7b\_v1\_full model, fine-tuned based on ranking data, the performance is not as high as the original Llama-2-7B model. This could be partly because the 9rank\_zephyr\_7b\_v1\_full was fine-tuned based on the MS dataset, and utilizes a listwise reranking optimizer, which differs from our setting. In summary, it appears that fine-tuned Llama2-based models may not perform as well as the original non-fine-tuned baseline models on specific tasks. However, as the model size increases, our method can leverage the capabilities of larger models to enhance the ranking performance.

\subsection{The Impact of Prompts}
\label{sec: prompt}

\begin{table}[t]
\centering
\resizebox{\linewidth}{!}{%
\begin{tabular}{c|c}
\toprule
Name & Content\\
\midrule
p1 & please generate a question for the input passage\\
p2 & please generate a Question for the input Passage\\
p3 & what is the question for the input passage\\
p4 & given the hints, please generate a question for the input passage\\
p5 & please generate a question for the input passage based on the hints\\
\bottomrule
\end{tabular}}
\caption{Different hard prompts utilized in UPR and Q-PEFT-based models.}
\label{tab: hard_prompt}
\end{table}

\begin{figure}[t]
\centering
\includegraphics[width=0.35\textwidth]{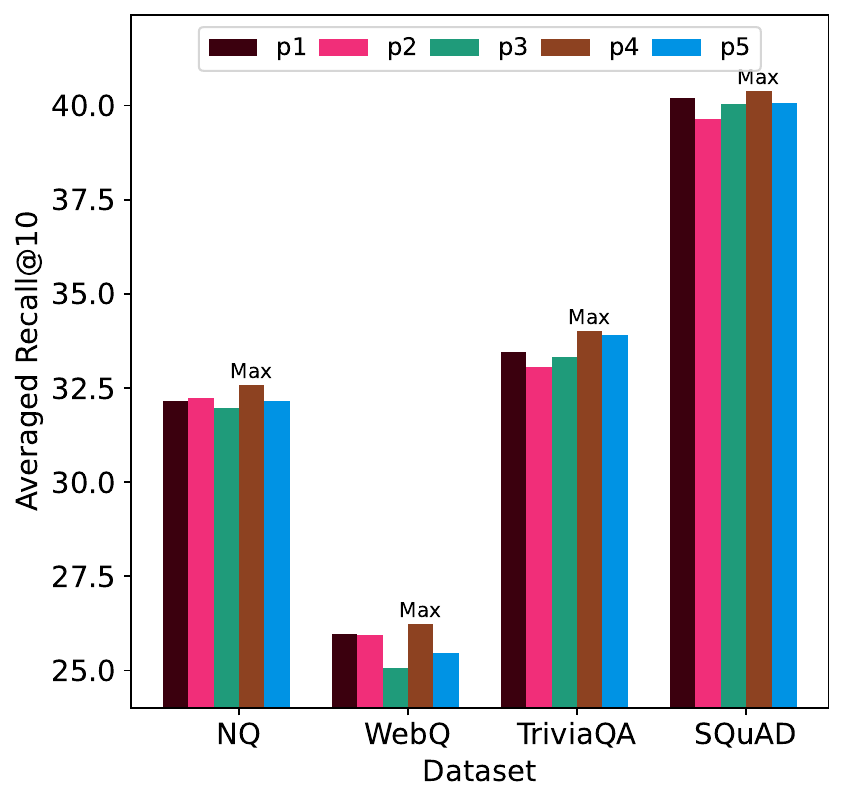}
\caption{Performance comparison of different hard prompts on various Datasets, the performance of Q-PEFT evaluated by the average Recall@10 Metric across various retrievers.}
\label{fig: upr_prompt}
\end{figure}

Prompts play a crucial role in Large Language Models (LLMs), guiding them to generate the expected output for various tasks. In our experiments, to ensure optimal results for both our proposed BE models and the baseline models, and to verify the stability of these results, we manually defined five task-related prompts. Table \ref{tab: hard_prompt} lists the prompts used in our experiments. We observe that prompts p4 and p5 are specifically designed to encourage the LLM to focus more on document hints during generation than prompts, compared to prompts p1, p2, and p3. Utilizing these five sets of prompts, we conducted a comparative analysis of the experimental outcomes of the baseline models and our proposed models across various datasets and retrievers. Surprisingly, the results show that prompt p4 consistently outperformed the others in different datasets, irrespective of changes in the baseline model or retriever. Although the improvement of p4 over other prompts is relatively minor, the consistent effectiveness across various experimental settings suggests that prompts with hints can indeed contribute to stable improvements in different models. In Figure \ref{fig: upr_prompt}, we compared the performance of different prompts across datasets, where the heights of the bars represent the average Recall@10 metric results of our proposed method based on different retrievers. Notably, prompt p4 demonstrated consistent results across various datasets. We employed prompt p4 in Table \ref{tab: main}, which effectively reduced variability in model outcomes attributable to differing prompts while maintaining optimal performance.


\subsection{The Impact of Training Size}
\label{sec: training_size}

\begin{figure}[!htbp]
    \centering
    \begin{subfigure}[b]{0.46\textwidth}
        \includegraphics[width=\textwidth]{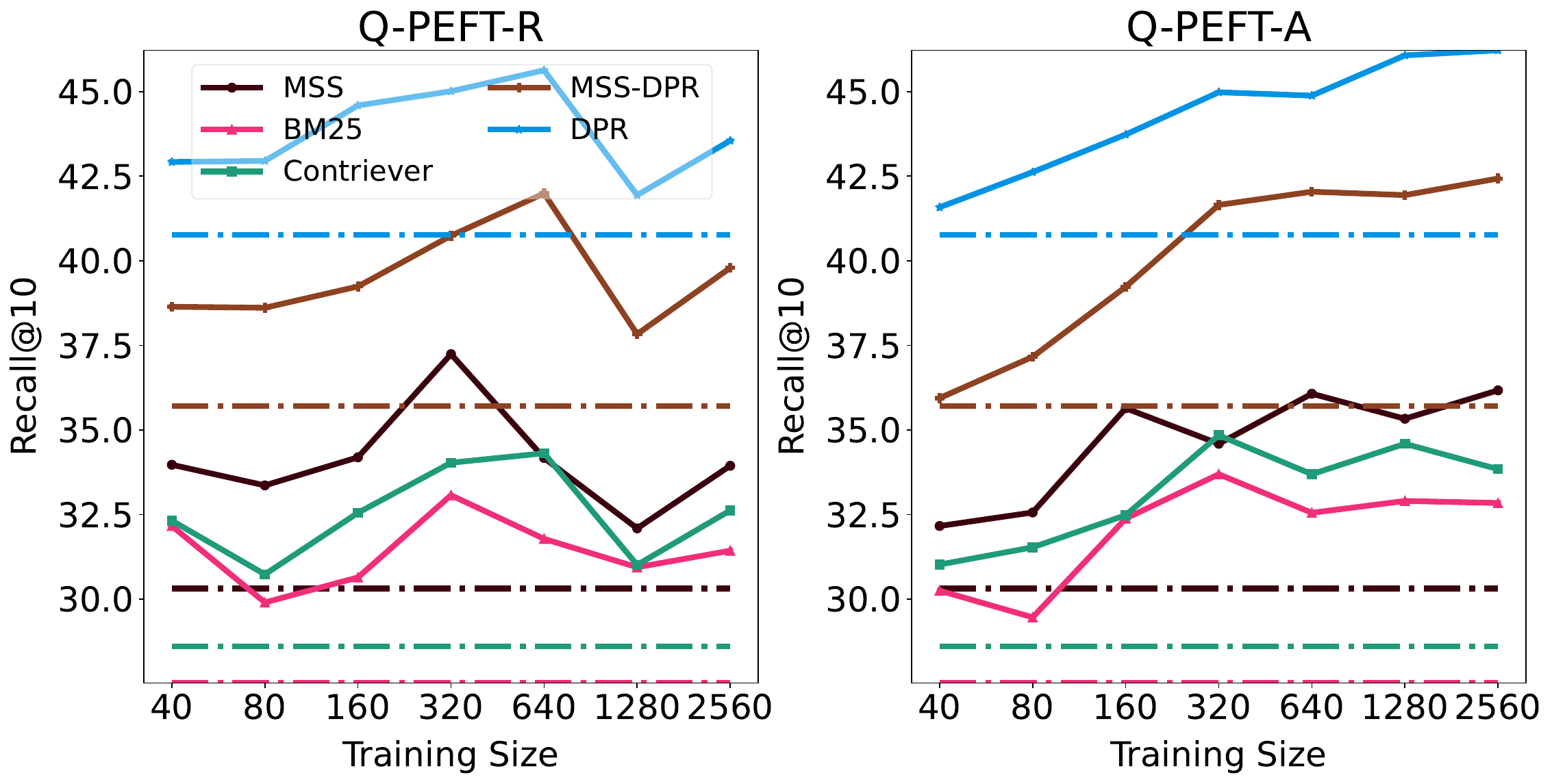}
        \caption{NQ}
        \label{fig:nq}
    \end{subfigure}
    
    \begin{subfigure}[b]{0.46\textwidth}
        \includegraphics[width=\textwidth]{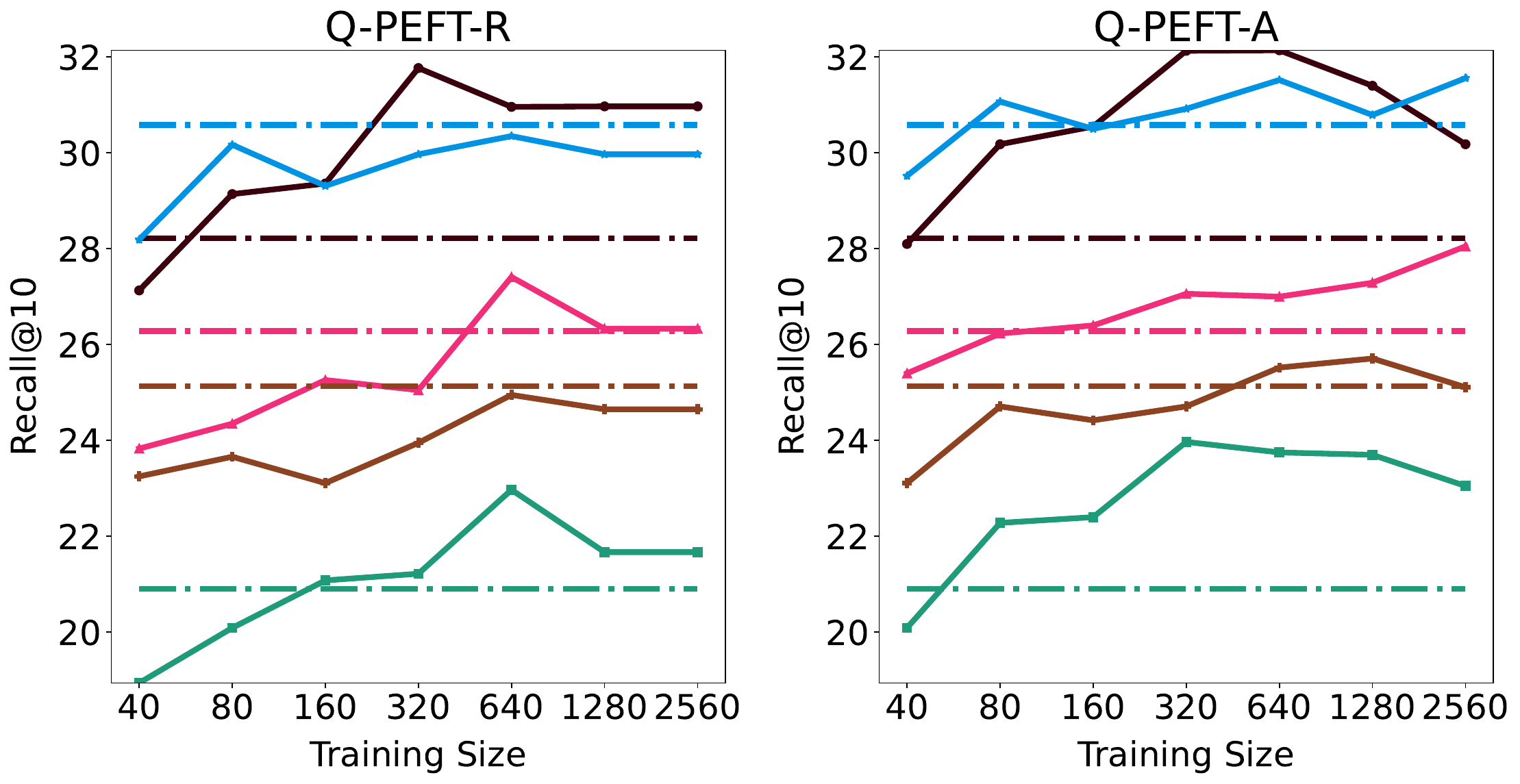}
        \caption{WebQ}
        \label{fig:webq}
    \end{subfigure}
    

    \caption{Comparative analysis of Recall@10 on NQ and Webq dataset for UPR and Q-PEFT-based models across various retrievers and different training sizes ($k$).}
    \label{fig: training_size}
\end{figure}

With efficient parameter tuning, training size has emerged as an important factor in our training process. We explored the impact of different training sizes on our proposed Q-PEFT model. Figure \ref{fig: training_size} primarily demonstrates the performance of our proposed Q-PEFT algorithm in the Recall@10 metric on the NQ and Webq datasets, based on different retrievers. The dashed line represents the UPR baseline model performance. As the UPR baseline is a non-trainable model, its results depend solely on the retriever used, regardless of training size changes. Overall, our proposed Q-PEFT models show a significant improvement over the UPR baseline. Comparing the Q-PEFT-R and Q-PEFT-A algorithms, the former exhibits less stability in Recall@10, with larger fluctuations based on different training sizes. On the NQ dataset, the Q-PEFT's performanQ-PEFTinitially increases with training size, then decreases, and subsequently increases again, showing significant variability. In contrast, the performance trend of the Q-PEFT algorithm on the Webq dataset is different, showing steady growth with increasing training size and stabilizing after reaching its peak. Compared to Q-PEFT-R, the Q-PEFT-A algorithm demonstrates smaller fluctuations in both the NQ and Webq datasets. The growth curve is smoother, increasing with training size, albeit more slowly than Q-PEFT-R, and generally achieving higher peak performances. The observed results can be attributed that, in the Q-PEFT-R algorithm, the model is able to more directly extract words in the document that have a significant impact on the reranking task. On the other hand, the Q-PEFT-A mainly leverages the attention weights between the query and the document, utilizing more implicit information within the semantic context. Consequently, while the improvement during training is slower, it demonstrates more robust performance in terms of semantics.

\begin{figure}[htbp]
    \centering
    \begin{minipage}{0.23\textwidth}
        \begin{subfigure}[b]{\textwidth}
            \includegraphics[width=\textwidth]{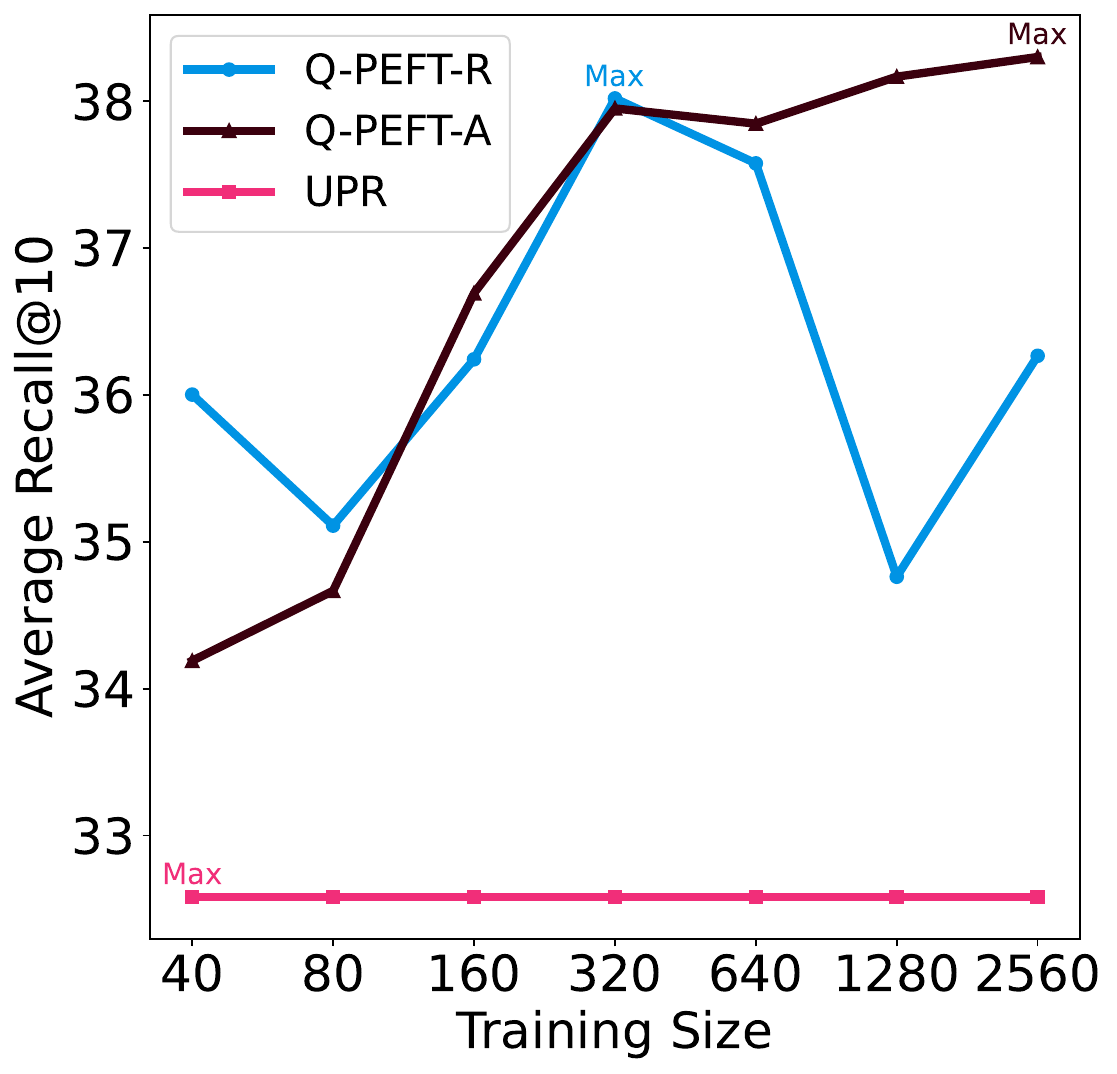}
            \caption{NQ}
            \label{fig:nq}
        \end{subfigure}
        
        \begin{subfigure}[b]{\textwidth}
            \includegraphics[width=\textwidth]{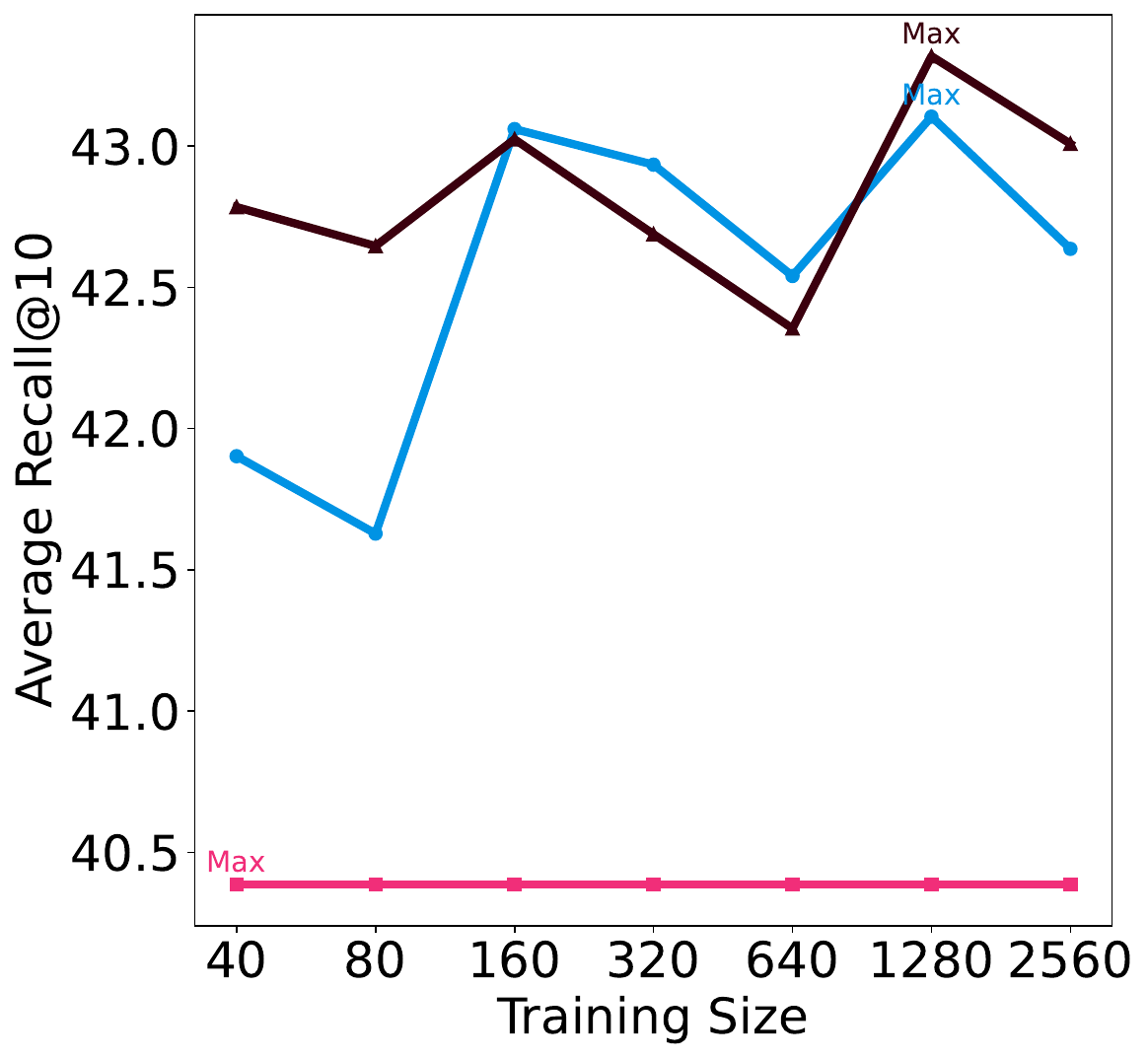}
            \caption{WebQ}
            \label{fig:webq}
        \end{subfigure}
    \end{minipage}
    \begin{minipage}{0.23\textwidth}
        \begin{subfigure}[b]{\textwidth}
            \includegraphics[width=\textwidth]{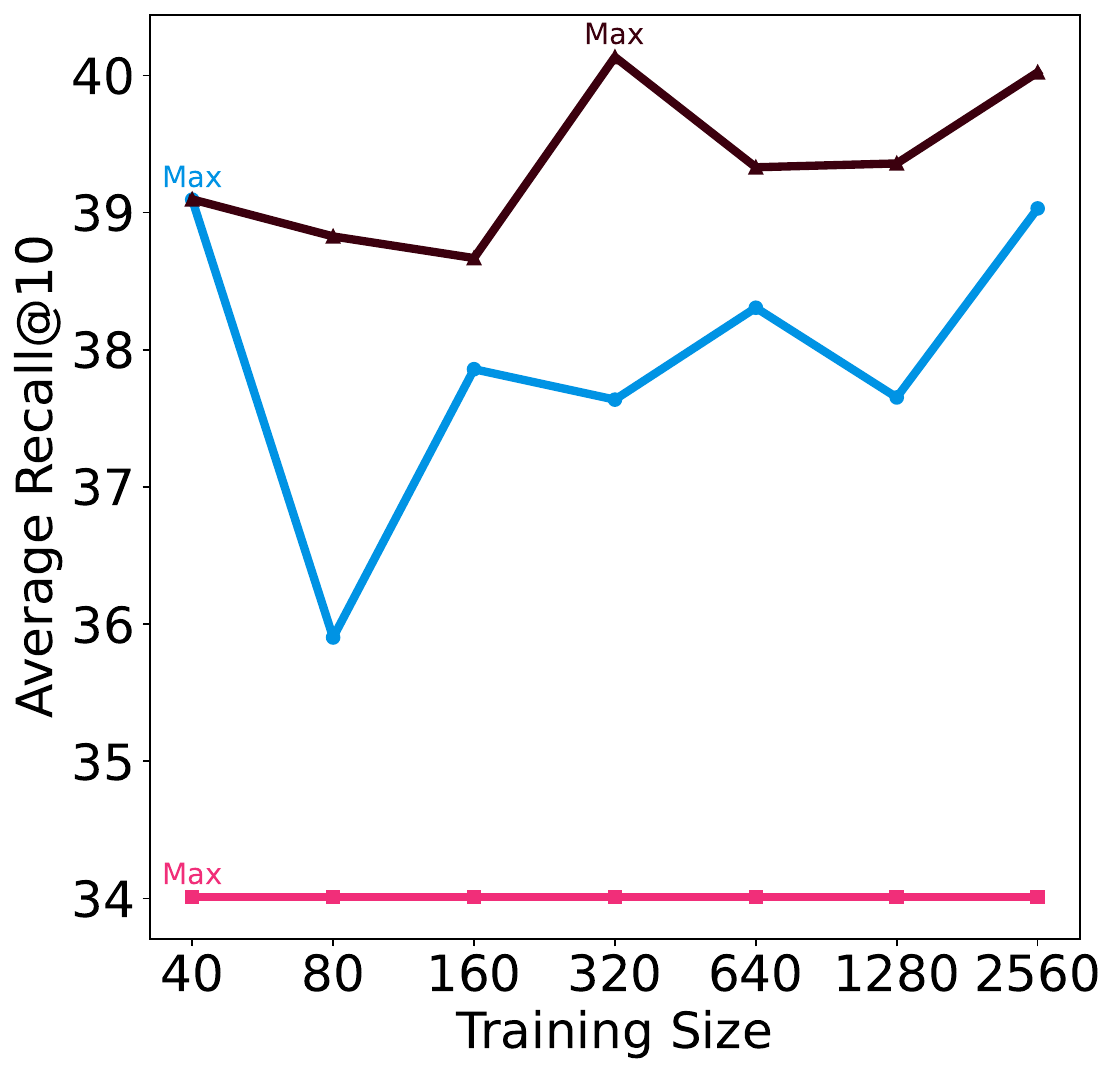}
            \caption{TriviaQA}
            \label{fig:trivia}
        \end{subfigure}

        \begin{subfigure}[b]{\textwidth}
            \includegraphics[width=\textwidth]{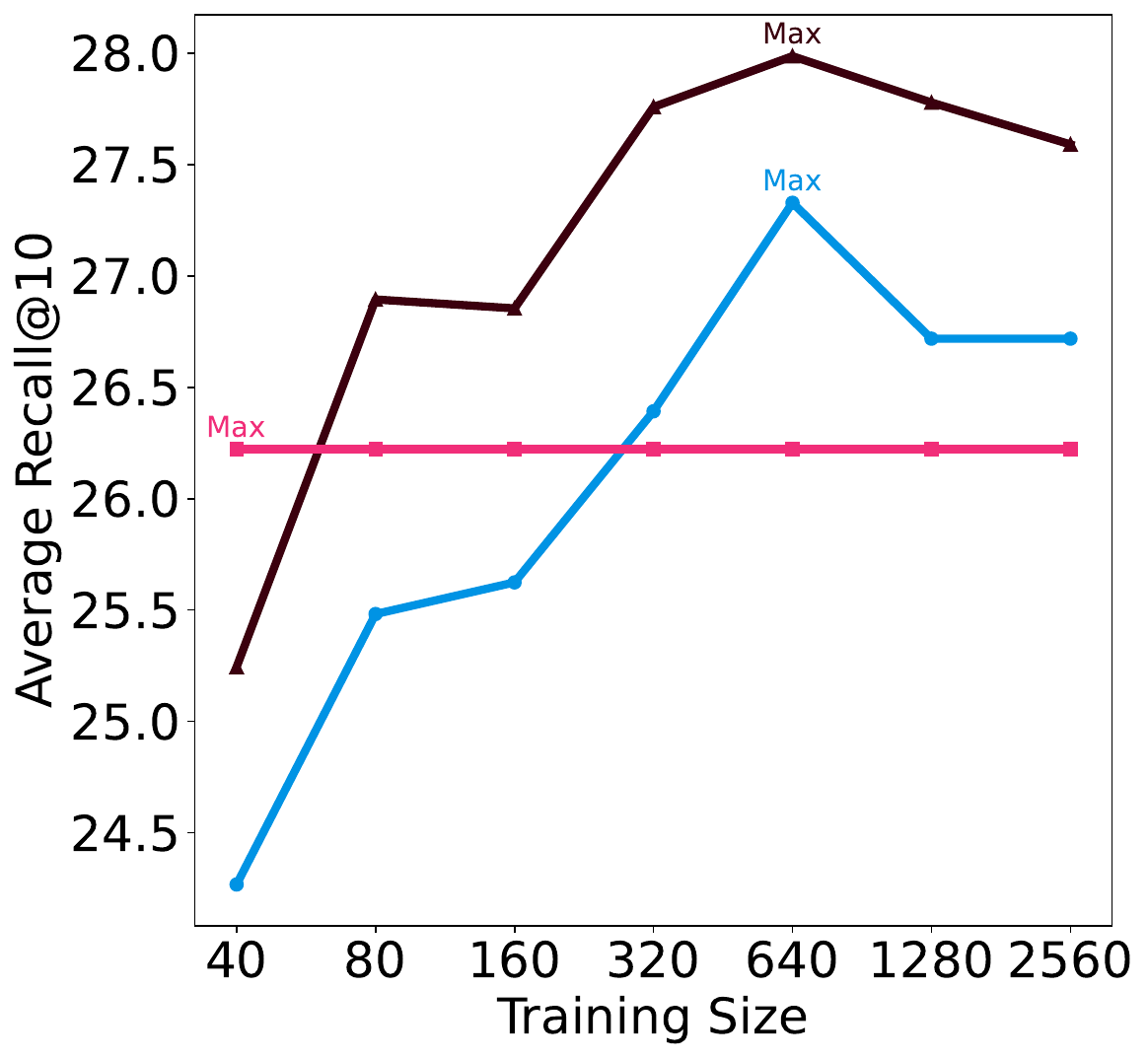}
            \caption{SQuAD}
            \label{fig:squad1}
        \end{subfigure}
    \end{minipage}
    \caption{Comparative analysis of the average Recall@10 for five retrievers using UPR and Q-PEFT-based models across various training sizes ($k$), evaluated respectively on the NQ, Webq, Trivia, and Squad1 datasets.}
    \label{fig: training_size_average}
\end{figure}

This trend is also reflected in Figure \ref{fig: training_size_average}. The x-axis in Figure \ref{fig: training_size_average} represents training size, while the y-axis shows the average Recall@10 across different retrievers. Different colored lines represent different models. The curve changes in Figure \ref{fig:nq} and Figure \ref{fig:webq} illustrate the performance trends of our proposed model with varying training sizes, corresponding to the results in Figure \ref{fig: training_size}. The results in Figure \ref{fig:trivia} and Figure \ref{fig:squad1} validate the stability of our proposed model, achieving significant results across various datasets. In Table \ref{tab: main}, the inference results are based on the optimal training size for each model and retriever.


\section{Conclusion and Future Work}

In this work, we propose a novel query-dependent parameter-efficient fine-tuning (Q-PEFT) approach for text reranking. We freeze the original parameters of the LLMs and introduce a new query-dependent module. Our method distinctively uses the query to extract the top-$k$ tokens from concatenated documents as contextual clues. Further, we augment Q-PEFT-A by replacing the retrieval mechanism with a multi-head attention layer, which guides the LLMs to generate more document-specific synthetic queries, thus influencing the ranking relevance score between the query and documents. Extensive experiments conducted on four datasets have demonstrated the efficiency of our proposed approach in ranking tasks. For future work, we hope to combine our query-dependent module with a soft prompt module, which could automatically adapt to different tasks without a hard prompt. Our query-dependent module is not only effective for passage reranking but can also be adapted to other ranking scenarios, making the exploration of these other scenarios an additional direction for research.

\begin{acks}
\end{acks}

\bibliographystyle{ACM-Reference-Format}
\bibliography{sample-base}

\appendix

\end{document}